\documentclass[10pt,twocolumn,letterpaper]{article}

\usepackage{cvpr}
\usepackage{times}
\usepackage{epsfig}
\usepackage{graphicx}
\usepackage{amsmath}
\usepackage{amssymb}
\usepackage{tabularx}

\graphicspath{{Images/}}

\usepackage[breaklinks=true,bookmarks=false]{hyperref}

\cvprfinalcopy 

\def\cvprPaperID{****} 

\begin{document}

\title{Modeling Urban Food Insecurity with Google Street View Images}

\author{David Li\\
Stanford University\\
{\tt\small davidwl@stanford.edu}
}

\maketitle

\begin{abstract}
Food insecurity is a significant social and public health issue that plagues many urban metropolitan areas around the world. Existing approaches to identifying food insecurity rely primarily on qualitative and quantitative survey data, which is difficult to scale. This project seeks to explore the effectiveness of using street-level images in modeling food insecurity at the census tract level. To do so, we propose a two-step process of feature extraction and gated attention for image aggregation. We evaluate the effectiveness of our model by comparing against other model architectures, interpreting our learned weights, and performing a case study. While our model falls slightly short in terms of its predictive power, we believe our approach still has the potential to supplement existing methods of identifying food insecurity for urban planners and policymakers. 
\end{abstract}

\section{Introduction}

Food insecurity is defined as a household-level economic and social condition of limited or uncertain access to adequate food \cite{USDAfoodinsecuritydef}. This condition affects millions of people globally and poses a significant public health concern for policymakers. Detecting households and neighborhoods experiencing food insecurity is crucial in both investigating social inequality and helping policymakers and other organizations make informed decisions about where to best allocate food resources. Traditional approaches to detecting food insecurity rely on collecting survey and demographic information from non-profit or government organizations. Data produced by these methods, however, is expensive and temporally-limited, often resulting in out-dated information. Furthermore, these datasets can suffer from spatial sparsity and may miss vulnerable populations who are underrepresented in survey responses.

To overcome these limitations, there is a growing interest in exploring alternative, data-driven methods for identifying food insecurity at scale. Previous research has considered how machine learning models applied to structured data—such as satellite imagery-derived land use features, census data, housing conditions, and socioeconomic indicators—can be used to estimate food insecurity across geographic regions \cite{A4} \cite{NICAAVRAM2021469} \cite{FoodInsecurityRemoteSensing}. These approaches have shown promise in improving temporal granularity and identifying food insecurity with consistent accuracy. 

However, to our knowledge, no study has been done on the effectiveness of street-level image data in modeling food insecurity. We believe this project is interesting for two reasons. First, street-level imaging is the most granular visual data we have regular access to within the realm of urban planning. Compared to traditional methods of identifying food insecurity from satellite imaging, street-level imaging is unique in the spatial information it provides. Second, we believe the relevance of street-level images will continue to increase in importance and availability as hopes for autonomous driving continue to grow. Therefore, being able to harness such rich data for important urban problems like food insecurity will likely also prove fruitful. 

This project seeks to fill this literature gap by investigating whether physical features of the built-environment, as captured through Google Street View, can be used to predict food insecurity at the census tract level in urban metropolitan areas. We pull street-view images from 25 of the biggest U.S. cities and use them to fine-tune a ResNet-18 model pretrained on Places365. We then propose a mechanism for aggregating individual images to predict food insecurity on a neighborhood level. Overall, we find that our best model is able to achieve an accuracy rate of 0.80 and an F1-score of 0.71. We also find that our model tends to rely much on the presence of well-maintained housing or other building structures when making predictions. Though this accuracy is not state-of-the-art, we believe our model holds the potential to supplement existing methods of identifying food insecurity for policymakers and researchers. 

It is important to note here that we recognize the limitations of our methodology in this task. Google Street View images that may consist of many noisy data points. It could be that street view images provide no meaningful information for predicting food security. Therefore, the purpose of this study is more exploratory rather than prescriptive. 

\section{Related Work}

\noindent \emph{A. Modeling Food Insecurity}

Existing approaches to model and analyze food insecurity have primarily taken one of two approaches. First, a qualitative approach like found in ~\cite{A3}. The qualitative approach relies primarily on survey data gathered from willing participants. These surveys primarily ask questions like "Did your family have enough food during the past 30 days" or "Do you have easy access to a grocery store?" Questions like these are directly helpful in allowing policymakers to pinpoint specific households or neighborhoods experiencing food insecurity. Other studies like ~\cite{A4}, however, have taken a much more data-driven quantitative approach to modeling food insecurity. The quantitative approach relies primarily on using demographic, agricultural, urban, and socioeconomic data to model and even predict food insecurity hot spots at scale. Some studies like \cite{NICAAVRAM2021469} have shown how machine learning can also be applied to predicting food insecurity through food sharing behaviors in the UK. Some work has also been done in applying deep learning computer vision techniques to identifying food insecurity. Studies like \cite{FoodInsecurityRemoteSensing} have looked at how remote sensing and satellite imaging can predict crop-yield. \\

\noindent \emph{B. Street-level Image Data}

Street-level images, and specifically Google Street View images, have been used in a variety of computer vision applications. One of the first studies to popularize the use of street-level data was ~\cite{A1} in 2017. ~\cite{A1} focused primarily on applying deep learning techniques to predict the socioeconomic and demographic attributes based on Google Street View data alone. The model showed significant promise, relying primarily on a unique association between cars and people to make predictions. The success of the study exemplified the potential of street-level data to provide both unique and specific information about the built-environment in ways that previous data sources like satellite imaging could never do.

As a result, the use of street-level data in deep learning research has rapidly increased. Previous studies like ~\cite{A6}, ~\cite{A7}, and ~\cite{chen_artificial_2024} have focused primarily on predicting health and well-being outcomes by analyzing the quality of infrastructure in the built environment. Other studies like ~\cite{A2} and ~\cite{huang2024citypulsefinegrainedassessmenturban} look at how Google Street View images can be used to understand gentrification and physical urban change over time.

However, a gap exists in this line of research in understanding the relationship between the built-environment and food insecurity. This project attempts to fill this gap by applying deep learning and street-level imaging to identify visual patterns in neighborhood infrastructure that may serve as features for identifying food insecurity at the census tract level. Our methods are largely inspired by the work done in \cite{A2} and adapted for the task of mapping food insecurity. By extending prior work on street-level deep learning models, this study introduces a novel application of computer vision. 

\section{Data}

\textbf{Google Street View.} We collected a total of approximately 25,000 Google Street View images across 25 of the largest urban metropolitan areas in the United States (about 1000 images per city). To select the geo-spatial coordinates for each image, we randomly sample locations along the road network in each urban area with a buffer to reduce overlap. 

\begin{figure}
    \centering
    \includegraphics[width=2.7cm]{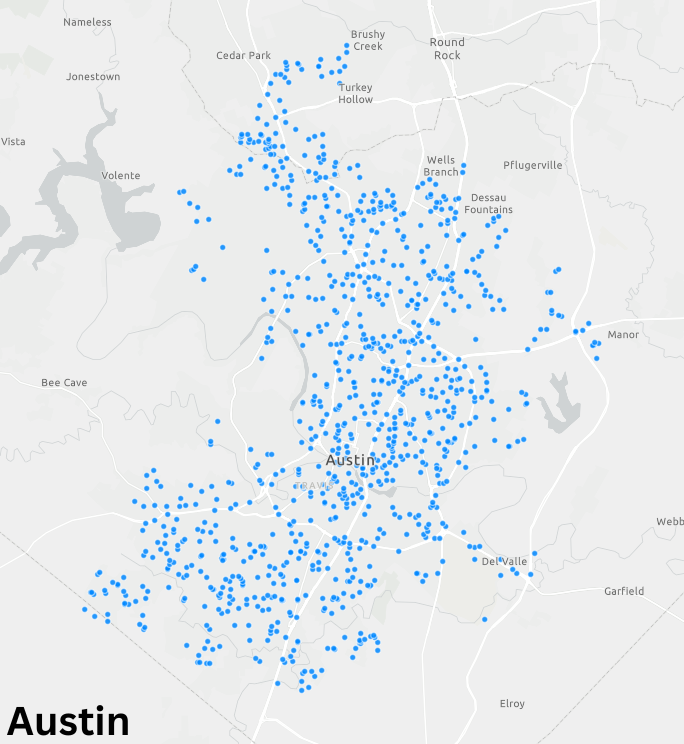}
    \includegraphics[width=2.7cm]{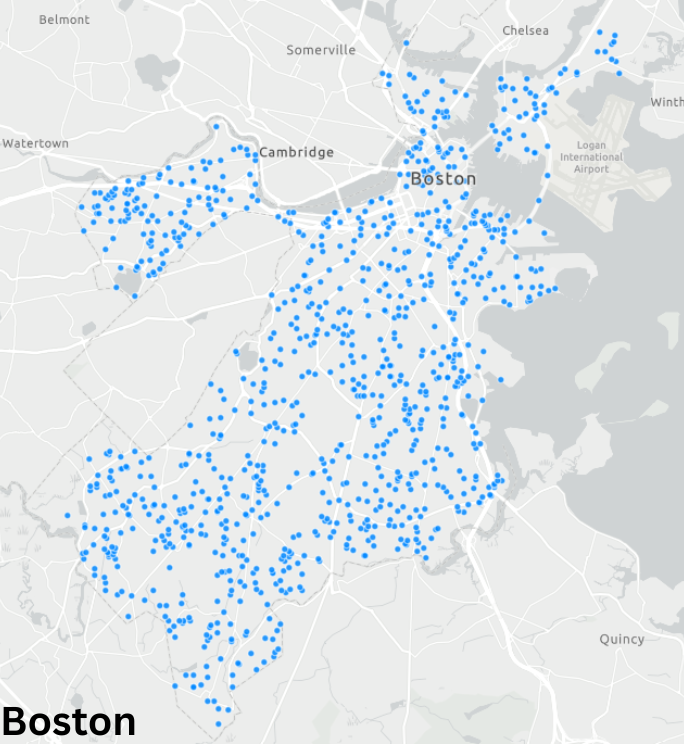}
    \includegraphics[width=2.7cm]{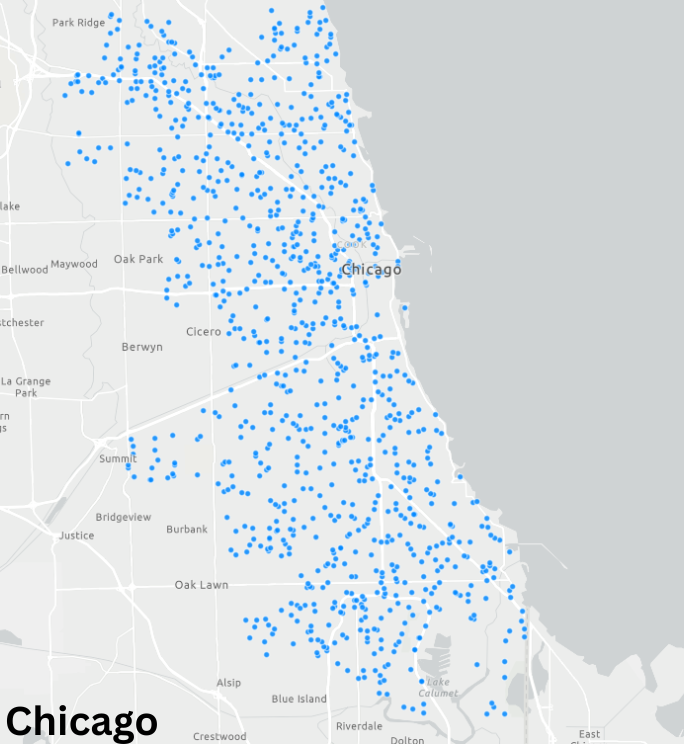}
    \caption{Distribution of sampled street views in three studied cities.}
    \label{fig:sampled_data}
\end{figure}

\textbf{Food Insecurity.} Food insecurity data is extracted from the USDA's 2019 Food Access Research Atlas \cite{usdafoodatlas}. Each row of the dataset includes information on food accessibility for a given census tract within the United States. The Food Access Research Atlas assigns any of four food insecurity labels to a census tract depending on how far food insecurity exists from that tract. If no label is assigned, the tract is considered food secure. However, because of the spatial limitations of street view data, we choose to ignore the distance measurement and instead define a census tract to be food insecure if any one of the four food insecure labels is assigned to that tract, thus creating a binary representation of food insecurity. 

\textbf{Data Labeling.} We use the food insecurity data from the Food Access Research Atlas to label each image in our dataset. Since food insecurity is defined as a binary label for a census tract, each image is first mapped to its corresponding census tract based on the U.S. Census Bureau's 2010 census tract boundaries (consistent with the Food Access Research Atlas data). We then label each image as either food insecure or food secure, depending on the label of its parent census tract. 

\textbf{Data Augmentation.} We also choose a few standardized pre-processing techniques to improve the learnability of our model. We first re-size all images to be 256x256 pixels before applying a random crop of 224x224 in order to fit within the dimension constraints of our model architecture. We then also apply a color jitter to each image to prevent the model from learning patterns in brightness and contrast, an issue particularly relevant for street-view images because of differences in sunlight and regional climate. Lastly, we apply a normalization to our images to stabilize the training process. 

\section{Method}

\subsection{Problem Statement}

This project aims to develop a computer vision-based method to model food insecurity using only publicly available Google Street View images. The input to the model is a set of geolocated Street View images sampled across U.S. urban areas, each associated with a census tract. The output is a binary classification indicating whether the corresponding census tract is food insecure (1) or not (0), based on existing food insecurity labels derived from USDA data. 

Because our input and output differ in terms of spatial granularity, we propose the following definition for neighborhood aggregation. An urban metropolitan area consists of $N=\{n_1,n_2,...,n_N\}$ different census tracts. Each census tract $n_j$ (which we also refer to as a neighborhood), is composed of K street-level images such that $n_j=\{s_j^{(1)},s_j^{(2)},...,s_j^{(K)}\}$. 

In accomplishing our aforementioned task, we propose the following steps: 

\begin{itemize}
  \item Step 1: Extract relevant features that may model food insecurity within individual images (e.g. neighborhood condition, store fronts, etc.).
  \item Step 2: Aggregate individual images to the census tract level and predict the neighborhood's food insecurity status. 
\end{itemize}

\noindent \emph{Step 1. Feature Extraction}

To identify food insecurity among urban street-level images, our first step is to extract relevant features that may prove useful for modeling the physical appearances of food insecurity. We choose to utilize an existing ResNet-18 model pre-trained on the Places365 dataset to extract high-level semantic meanings from each image. 

ResNet-18 is a deep convolutional neural network architecture comprising 18 layers, designed to facilitate the training of very deep networks by introducing residual connections. This design allows for the extraction of rich feature representations, making it suitable for our task. 

We choose to pre-train our model on Places365 (\cite{A5}), a large-scale scene recognition dataset comprising approximately 10 million images categorized into 434 distinct scene classes, such as "urban street," "residential neighborhood," and "supermarket." In the context of identifying food insecurity through street-level imagery, Places365 provides the model a way to identify features pertinent to the urban environment.

\noindent \emph{Step 2. Neighborhood Aggregation}

A neighborhood $n_j$ is composed of $K$ different street-level images, and there exists a single binary label $Y$ indicating whether or not the neighborhood experiences food insecurity. By assuming no order and dependency between all $K$ images in a neighborhood, we propose a gated attention mechanism for aggregation as first discussed in \cite{ilse2018attentionbaseddeepmultipleinstance}. Let $H=\{h_1,h_2,...,h_k\}$ represent the bag of $K$ embeddings extracted from each individual image. We can then calculate the neighborhood feature vector with a weighted average of each instance embedding: 
\begin{equation}
n_j= \sum_{k=1}^{K} a_k h_k
\end{equation}

\noindent such that: 
\begin{equation}
a_k = \frac{
    \exp\left\{
        \mathbf{w}^\top \left( \tanh(\mathbf{V} \mathbf{h}_k^\top) \odot \sigma(\mathbf{U} \mathbf{h}_k^\top) \right)
    \right\}
}{
    \sum_{j=1}^{K} \exp\left\{
        \mathbf{w}^\top \left( \tanh(\mathbf{V} \mathbf{h}_j^\top) \odot \sigma(\mathbf{U} \mathbf{h}_j^\top) \right)
    \right\}
}
\end{equation}

\noindent where $w \in \mathbb{R}^{L \times 1}$ and $V,W \in \mathbb{R}^{L \times M}$ are learnable parameters. The application of such a weighting mechanism is motivated by two reasons. First, it is important to recognize that not all images in a neighborhood contribute equally in identifying food insecurity. For instance, an image of a grocery store or shopping center will tend to be much more informative than an image of a highway or road. If we simply performed a mean pooling on all images in a neighborhood, this could dilute the overall feature representation. Second, because of the way our data was extracted, the number of images per census tract varies across neighborhoods, making an attention-based weighting mechanism necessary for handling this imbalance. 

Lastly, we feed our pooled feature vectors into a fully connected layer to get our final tract-level prediction: 
\begin{equation}
\hat{y}_j = w^Tn_j+b_j
\end{equation}

Rather than using traditional cross-entropy loss to train our model, however, we opt for a weighted cross-entropy loss in order to account for class imbalance (there are many more instances of food secure neighborhoods than food insecure neighborhoods). Thus, we have $y_j=1$ when the neighborhood is food insecure and $y_j=0$ when the neighborhood is food secure and 
\begin{equation}
L(n_j) = -w [y_j \log \sigma (\hat{y_j}) + (1-y_j) \log(1 - \sigma(\hat{y_j}))]
\end{equation}

\noindent where the positive class weight is $w=\frac{N_0}{N_1}$ and $N_1$ and $N_0$ are the number of positive (food insecure) and negative (food secure) examples in the training set. 

We then use this loss function to back-propagate through our model, fine-tuning the weights to suit our task. It's important to note that we choose to train our loss function on the census tract level. This allows our gated attention model to learn, at a high-level, how to distinguish between important and unimportant images in a tract. 

\subsection{Alternative Methods}
\noindent \emph{Alternative I: Baseline}

For our baseline, we adopt a simplified version of our two-step approach to classify food insecurity at the census tract level. First, we extract high-level features from each street-view image using a ResNet-18 model pre-trained on Places365. We then group the feature embeddings by their corresponding census tracts and compute the average embedding for each tract. Lastly, we use the tract-level embeddings as inputs to a simple neural network classifier to predict the food security status of each tract. 

\noindent \emph{Alternative II: ViT}

We also consider the effectiveness of vision transformers for this task. Specifically, we use ViT-B/16, a model that takes in inputs of 224x224 pixels and splits them into patches of size 16x16 and processes them with multi-head attention. 

\noindent \emph{Alternative III: Larger ResNet}

We also consider the effectiveness of using a larger model like ResNet-50. ResNet-50 follows the same architectural structure as ResNet-18, except with 50 layers of convolution. 

\section{Experiments and Results}

\subsection{Predicting Food Insecurity}
For each of the architectures detailed above (except for the baseline), we train and evaluate the model as described in Steps 1 and 2. Our model is evaluated using accuracy and F1-score as core metrics. Accuracy is simply the number of predictions guessed correctly in the test set. F1-score is given as the following equation where $TP$ is the number of tracts correctly predicted as food insecure, $FP$ is the number of tracts incorrectly predicted as food insecure, and $FN$ is the number of tracts incorrectly predicted as food secure: 
\begin{equation}
F1=\frac{TP}{TP+\frac{1}{2}(FP+FN)}
\end{equation}

Lastly, we split our dataset into a training (60\%), validation (20\%), and test set (20\%) based on census tracts. To ensure the validation and test set look similar to the training set, we perform a stratified split according to the overall dataset's class imbalance. 

\noindent{\emph{Baseline}}

Our baseline method performed below expectations with a total validation accuracy of 0.74. The model achieves an F1-score of 0.81 for class 0 labels (food secure census tracts) and 0.42 for class 1 labels (food insecure tracts). This discrepancy indicates that the class imbalance in the dataset poses a strong barrier for the model. Overall, however, the model seems unable to extract relevant features for the task at hand. Even simply predicting the majority class (because of the imbalanced data) yields an accuracy of 0.72. 

\noindent{\emph{ViT Base}}

The ViT-B/16 was found to be incompatible for this task. Initial accuracy rates using this model hovered around 0.73, almost no better than simply predicting the class majority. We believe the transformer-based model faces limitations on this task for two primary reasons. First, due to limitations with Google API, we lacked a large quantity of training data, not allowing the model to generalize properly. Second, because of the image patching, ViTs may tend to perform worse when it comes to identifying fine-grained local detail, which is critical in identifying food insecurity indicators like building conditions or store-fronts. 

\noindent{\emph{ResNet-50}}

The ResNet-50 model, though a much larger model with many more layers and parameters than ResNet-18, proved to fair worse on our task. Our model was only able to achieve an accuracy of 0.76, an F1-score of 0.80 for food secure tracts, and an F1-score of 0.52 for food insecure tracts. We believe the reason why ResNet-50 fails to perform well on this task is due to the model's tendency to overfit on training data. While this model could achieve a high training accuracy rate of around 0.90, it's validation accuracy remained consistently low. Even efforts at regularization like the addition of dropout was unable to help the model generalize well. 

\noindent{\emph{ResNet-18}}

\begin{figure}
    \centering
    \includegraphics[width=7cm]{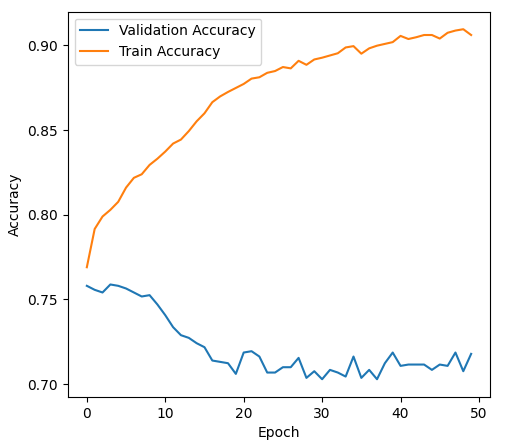}
    \caption{Training and validation accuracy plotted against training epochs before regularization.}
    \label{fig:pre-regularization accuracy}
\end{figure}

\begin{figure}
    \centering
    \includegraphics[width=7cm]{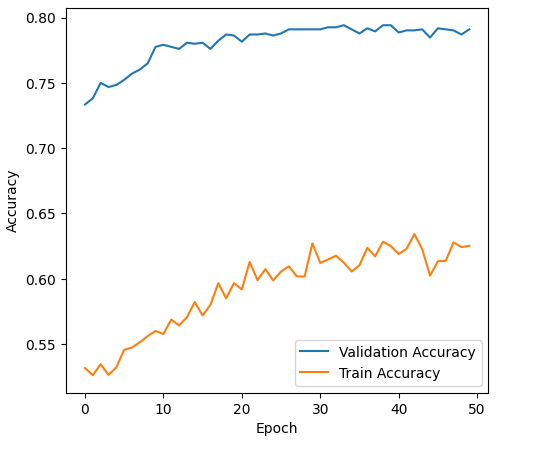}
    \caption{Training and validation accuracy plotted against training epochs after regularization.}
    \label{fig:post-regularization accuracy}
\end{figure}

We were able to achieve our best results with ResNet-18 as our backbone. Similar to ResNet-50, our model struggled initially with overfitting on training data (as seen in Figure \ref{fig:pre-regularization accuracy}). To mitigate this, we implemented weight decay, dropout, and label smoothing in our training. While these measures lowered the training accuracy significantly, the model was able to generalize better on the validation and test set (as seem in Figure \ref{fig:post-regularization accuracy}). 

We performed single-fold cross-validation on the following hyper-parameters: learning rate ($1e^{-5}$), weight decay ($1e^{-4}$), dropout rate ($0.9$), batch size ($64$), label smoothing ($0.1$), optimizer (Adam). We found that altering the dropout rate had the biggest impact on accuracy and F1-score, likely due to its strong regularization effect. 
\begin{table}[h]
\centering
\footnotesize
\begin{tabular}{|c|c|c|c|}
\hline
Learning Rate & Dropout Rate & Accuracy & F1-Score (Average) \\
\hline
1e-5 & 0.0 & 0.74 & 0.60 \\
1e-5 & 0.25 & 0.75 & 0.62 \\
1e-5 & 0.50 & 0.77 & 0.67 \\
1e-5 & 0.90 & 0.80 & 0.71 \\
\hline
\end{tabular}
\vspace{5pt}
\caption{Cross-validation results (some parameters not shown)}
\label{tab:cross-validation}
\end{table}

\subsection{Interpreting Learned Weights}

To uncover how our proposed method determines whether or not a census tract experiences food insecurity, we examine both our feature extraction and gated attention mechanism steps more in-depth. 

\noindent{\emph{Image Feature Extraction}}

\begin{figure}
    \centering
    \includegraphics[width=4cm]{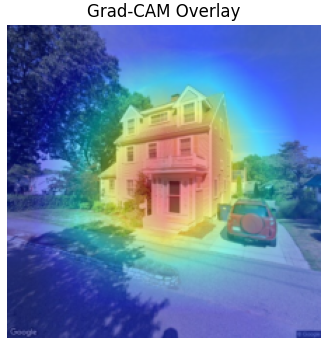}
    \includegraphics[width=4cm]{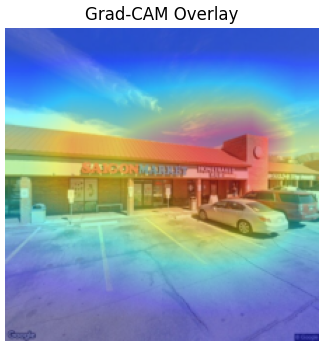}
    \caption{Good Grad-CAM heatmap of sampled street views.}
    \label{fig:grad-cam-good}
\end{figure}

Figure \ref{fig:grad-cam-good} shows good Grad-CAM visualizations that demonstrate how our model was, at times, able to successfully extract architecturally relevant features from our input images. The left image shows a concentrated activation around a well-maintained and modern house. Similarly, the right image shows strong activation around the commercial storefront of what appears to be a grocery store. These heatmaps highlight how the network is able to capture semantically meaningful spatial information like the presence of affluent housing or stores. 

However, we can also see examples of images (like in Figure \ref{fig:grad-cam-good}) where our model seemed unable to extract relevant features from the model. In the image on the left, we can see the model focusing primarily on parts of the road instead of the seemingly affluent apartment complex in the background. Similarly, in the image on the right, we can see the model focused on the sky and a few telephone poles. This indicates that our model at times does fail to extract meaningful information, which could contribute to it's subpar classification accuracy. 

\begin{figure}
    \centering
    \includegraphics[width=4cm]{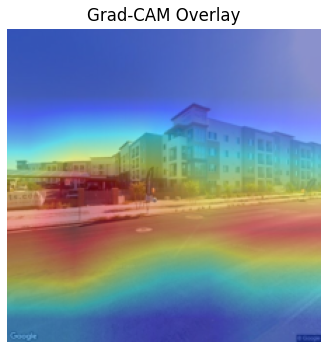}
    \includegraphics[width=4cm]{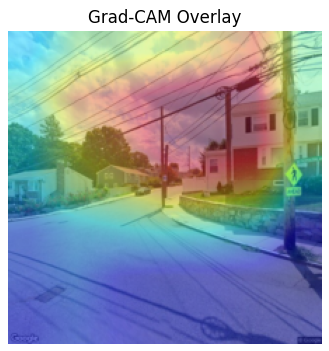}
    \caption{Bad Grad-CAM heatmap of sampled street views.}
    \label{fig:grad-cam-bad}
\end{figure}

\noindent{\emph{Gated Attention Weights}}

\begin{figure}
    \centering
    \includegraphics[width=2.7cm]{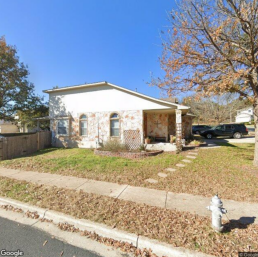}
    \includegraphics[width=2.7cm]{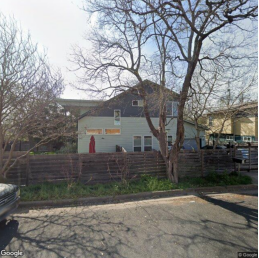}
    \includegraphics[width=2.7cm]{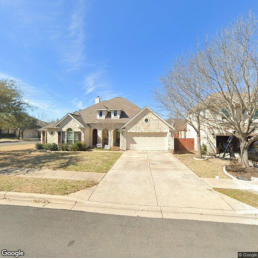}
    \caption{Examples of images given higher weights.}
    \label{fig:high_weights}
\end{figure}

\begin{figure}
    \centering
    \includegraphics[width=2.7cm]{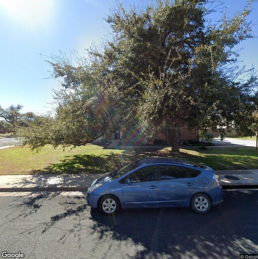}
    \includegraphics[width=2.7cm]{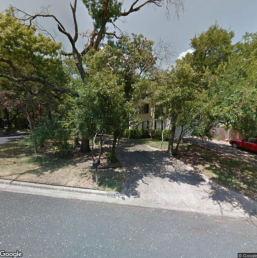}
    \includegraphics[width=2.7cm]{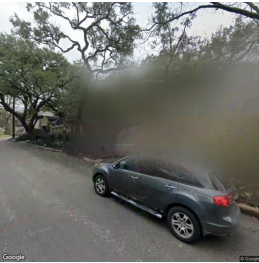}
    \caption{Examples of images given lower weights.}
    \label{fig:low_weights}
\end{figure}

To better understand how our model aggregates image features together, we examine the attention weights corresponding to individual images. The higher the attention weight, the higher the model will consider that image more influential in determining the overall prediction for a given census tract. Example images with higher weights and lower weights are shown in Figures \ref{fig:high_weights} and \ref{fig:low_weights}. In general, our model tends to assign higher weights to images with semantically meaningful structures like presence of well-maintained housing or other buildings while assigning lower weights to images that may lack such structures or images that are slightly blurred. 

\subsection{Case Study: Austin}

Given that our proposed method seeks to have practical applications in urban planning and public policymaking, we seek to apply our model to a specific city in question. For the sake of this case study, we choose to focus on Austin, Texas. In particular, we are interested in how well the model can generalize to a new city it has never seen before. To do this, we split our dataset into a training set with all images not in an Austin census tract (96\%) and a test set with all images taken in an Austin census tract (4\%). We then train our model following the same architecture as noted above. 

Unexpectedly, we discovered a significant drop in testing accuracy when predicting food insecurity in Austin census tracts with an accuracy rate of 0.70 and an F1-score of 0.76 for food secure tracts and 0.54 for food insecure tracts. This performance gap may be due to the model's limited ability to generalize visual features across different geographic regions. While our training data included images from a variety of urban environments, the visual and structural characteristics of Austin may have not been well represented. The model may have overfit to specific cues present in training cities that may not hold the same predictive value in Austin. This highlights a broader challenge when applying deep learning models on street-view imagery across different metropolitan contexts. 

\begin{figure}
    \centering
    \includegraphics[width=4cm]{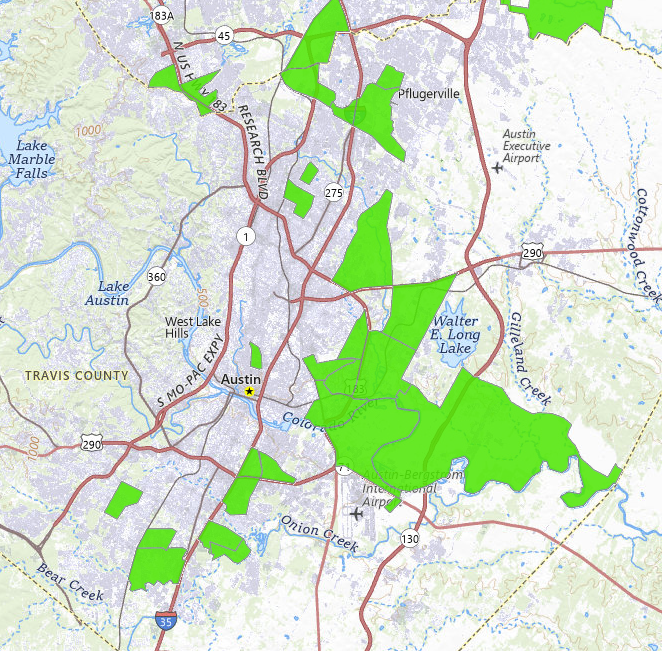}
    \includegraphics[width=4cm]{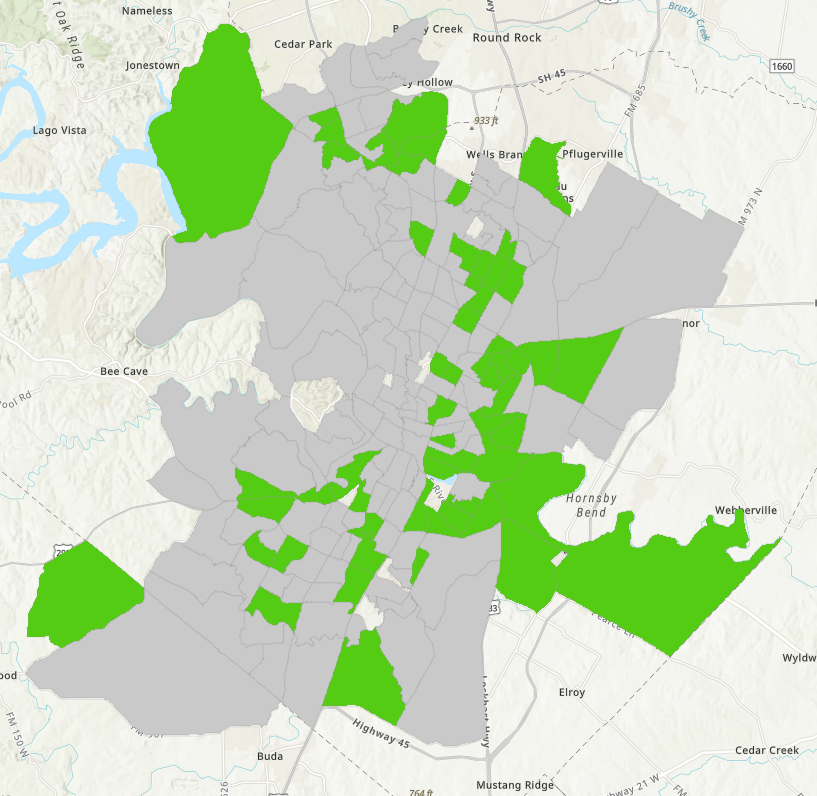}
    \caption{Side-by-side comparison of our model's predictions (right) and the USDA Food Atlas map (left).}
    \label{fig:austin-predictions}
\end{figure}

To better understand our model's predictions when compared to our ground-truth labeling (USDA Foot Atlas), we visualize them in Figure \ref{fig:austin-predictions}. In both images, census tracts colored green represent food insecure census tracts. Though our model does not necessarily identify food insecure tracts with complete precision, it tends to identify the general pattern or shape of food insecurity in the city. For example, our model successfully identified the large group of food insecure tracts in East Austin as well as the general location of tracts in the north and south parts of the city. This suggests that our model, though slightly lacking in accuracy, is still able to learn and reproduce spatial trends in food insecurity. 

\subsection{Exploring Multi-modal Data}

To begin to explore the use of other forms of data in our model, we will analyze the effect of including tabular data in our model's input. We choose to focus on median household income, a socioeconomic indicator that strongly correlates with food insecurity \cite{USDAfoodinsecuritydef}. To do so, we extract median household income data for each census tract based on the American Community Survey (ACS) 5-year survey in 2010 and slightly alter our model architecture accordingly. 

To incorporate this new data into our model, we take a late fusion approach by concatenating the output from our gated attention model with a normalized median household income value before passing it into our final linear classification layer.

\begin{figure}
    \centering
    \includegraphics[width=7cm]{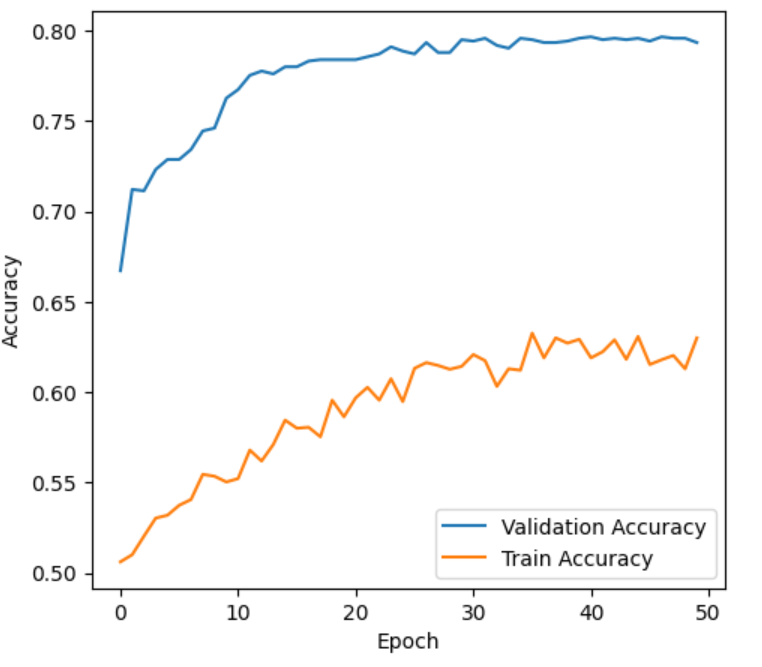}
    \caption{Training and validation accuracy for our multi-modal model plotted against training epochs after regularization.}
    \label{fig:post-regularization accuracy multimodal}
\end{figure}

Though our implementation is relatively naive, we expected our new model to perform slightly better in classifying food insecure tracts because of an additional layer of input data. However, we found no significant increase in accuracy or F1-score. After tuning hyper-parameters, we were still only able to achieve a test accuracy rate of 0.81 and an average F1-score of 0.71, an insignificant difference compared to our model that utilized only image data. The validation results, as shown in Figure \ref{fig:post-regularization accuracy multimodal}, don't differ much from the results previously mentioned in Figure \ref{fig:post-regularization accuracy}. 

We believe there are several possible reasons for a relatively insignificant performance increase. First, it could be that our naive integration of household income into our dataset may have resulted in the income feature being underutilized. Second, it also could be that our image data already encodes much of the same socioeconomic information reflected in median income, such as housing quality, infrastructure, and commercial presence, thus limiting the marginal benefit of adding income data. Future work would explore more sophisticated data types and methods of multi-modal integration. 

\section{Conclusion}

In this project we have explored a novel application of street-view images in detecting food insecurity in urban metropolitan areas. Previous computer vision studies using deep learning models in analyzing food insecurity have focused primarily on using survey data, tabular data, or high-granular visual data like satellite-based imaging. Our model utilizes a two-step approach. First, we extract relevant semantic features from individual images using a ResNet-18 model pretrained on Places365. Second, we aggregate image features together using a gated attention mechanism to make predictions at the census tract level. When compared to our baseline and alternative methods, our model tended to perform slightly better after strong regularization was applied. As a result, we believe that street-view images can provide meaningful information for deep learning models in predicting food insecurity. While our approach is not accurate enough to replace existing methods, we believe that it can be used to supplement models with a more granular view of the urban environment.

Our findings also testify to the potential of street-view imaging in urban planning research. As models become more sophisticated and data more abundant, integrating street-level imagery may offer a richer lens into the structural and environmental drivers of many urban issues. 

There are, however, some limitations with our approach in this project. First, the dataset we used was relatively small, containing only 25,000 images. This likely contributed to the problems we faced with overfitting in all of the model architectures we deployed. Another issue with our dataset is its tendency towards producing noisy or semantically meaningless information such as pictures of random highways, streets, or traffic intersections. Filtering out such low-quality data poses a significant challenge when attempting to use street-level imaging. Second, our labeling relied heavily on an over-simplified definition of food insecurity. As noted in the USDA Food Atlas, food insecurity is often a spatial and geographical problem (i.e. proximity to fresh and affordable food). Our analysis does not account for this spatial factor and assumes the problem of food insecurity is uniform across different neighborhoods. 

Future work could take one of two directions. First, while modeling food insecurity is an important task in and of itself, it may also be useful to explore building a time-step predictive model for food insecurity with street-view images. Being able to understand how food security changes overtime with neighborhood development may also be of use to policymakers and urban researchers. Second, future work could do more in attempting to integrate street-view imaging into existing food insecurity models. This may involve making use of existing census data, other demographic/socioeconomic indicators, or satellite imaging. 

\section{Contributions \& Acknowledgments}

All work was produced by the original author. There are no other contributors for this project.

{\small
\bibliographystyle{ieee}
\bibliography{egbib}
}

\end{document}